# XOR at a Single Vertex -- Artificial Dendrites

By John Robert Burger
Professor Emeritus
Department of Electrical and Computer Engineering
25686 Dahlin Road
Veneta, OR 97487
(jrburger1@gmail.com)

*Abstract* – New to neuroscience with implications for AI, the exclusive OR, or any other Boolean gate may be biologically accomplished within a single region where active dendrites merge.  This is demonstrated below using dynamic circuit analysis.  Medical knowledge aside, this observation points to the possibility of specially coated conductors to accomplish artificial dendrites.

**INTRODUCTION**

Boolean logic in an arborized biological neuron is an old topic that has been studied for quite some time (Stuart 2008; Poirazi 2003; Koch 1999; Segev 1995; Mel 1994).  The XOR in a historical neural network requires at least two layers of computations and three or more summing junctions each with a sigmoid model, as long ago proposed (Purves 2008; Squire 2008; Kandel 2000; Fromherz 1993, Zador 1992).  Surprising new results revealed below demonstrate that in spite of all these advances, not quite everything is known yet about the natural world of neurons.

Biologically, two pulses arriving simultaneously at a junction of active dendrites will annihilate when they collide (Fromherz 1993). Technically, pulses like this are *solitary waves* or *solitons*. They naturally occur within an idealized continuously active dendritic membrane (These and other waveforms come under the fuzzy term *action potential*).  Simulations show that within fairly broad limits, colliding solitons may compute the XOR at a dendritic vertex; any one pulse is transmitted but two are not. An important parameter is dendritic series resistance in the region of branching, since if it is too high, nothing is transmitted, and if it is too low, the common OR results. The possibility of a deterministic XOR for individual pulses simultaneously arriving at a dendritic vertex stands in stark contrast to and as a supplement to distributed networks of the past.

There are many equivalent ways to simulate dendritic solitons and series resistance.  The model below assumes only two types of ions, sodium and potassium.  It permits anyone with knowledge of circuit simulation to demonstrate the XOR or any other dendritic logic.  For circuit modeling purposes, sodium-induced current may be modeled macroscopically to be a current pulse with constant amplitude as in Fig. 1 (Top).  The pulse is triggered at a certain voltage $V_{TRIG}$.  As charge transfers through the membrane, sodium-induced current has been observed to shut off at a membrane potential of $V_{MAX}$.

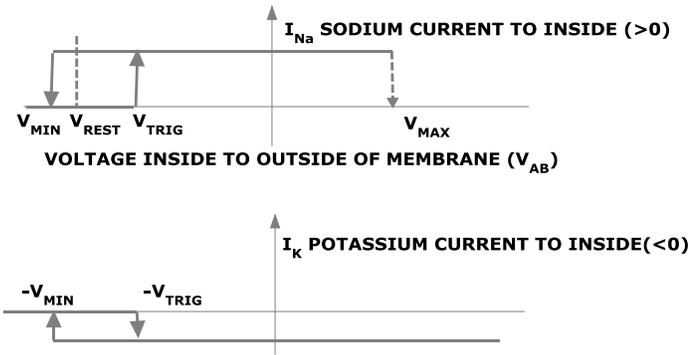

**Fig. 1 Simplified model of charges through a patch of active membrane**

In this the simplest of models, potassium-induced currents of lower magnitude are triggered at the same time to discharge the membrane as in Fig. 1 (Bottom). Potassium-induced currents continue to flow even when sodium currents are cut off; potassium currents are switched off only when voltage reaches $V_{MIN}$, after which the sensitive regions of the membrane recover to equilibrium. Although this model is simplified, the general shape of the resulting neural pulse is empirically correct.

Average sodium-induced current density may be set to be $J_{Na} = 269\ uA/cm^2$; potassium-induced current density may be set to be *60.8 uA/cm²* (Burger 2009). Assuming typical neural parameters ($c = 1\ uF/cm^2$, $g = 0.3\ mS/cm^2$, $\rho = 15.7\ Ohm\text{-}cm$) and a dendritic segment *500 um* long and *1 um* diameter, parameters maybe calculated as in Table 1.

**Table 1**
**Electrical Parameters for a Segment**

| | |
|---|---|
| C | 15.7 pF |
| $R_L$ | 212 M |
| R | 99.9 M |
| $I_{Na}$ | 4.22 nA |
| $I_K$ | 0.955 nA |
| $V_{REST}$ | -70 mV |
| $V_{TRIG}$ | -54 mV |
| $V_{MAX}$ | +48 mV |
| $V_{MIN}$ | -96 mV |

XOR Simulation – Simulation details are readily available elsewhere and will not be reproduced here (Burger 2008; 2009). Consider the merging of two active dendrites as in Fig. 2. All segments are assumed identical in this model.

To understand the conditions for Boolean logic, consider the circuit model in Fig. 3. A pulse from only one input $A_1$ soon arrives at Segment 6; this in turn activates Segment 7 and waveform $V_7(t)$. Segment 7 will, in turn, activate both Segments 8 and 16 creating pulses waveforms $V_8(t)$ and $V_{16}(t)$. Current sources in the above circuit are voltage dependent and not load dependent; $V_7(t)$ increases to a maximum of about +48 mV in this model without regard for the extra current used by the extra load segments. The result is a pulse that goes forward to the output $A_N$ while another pulse goes back to the input $A_{11}$ where it terminates because of the open circuit.

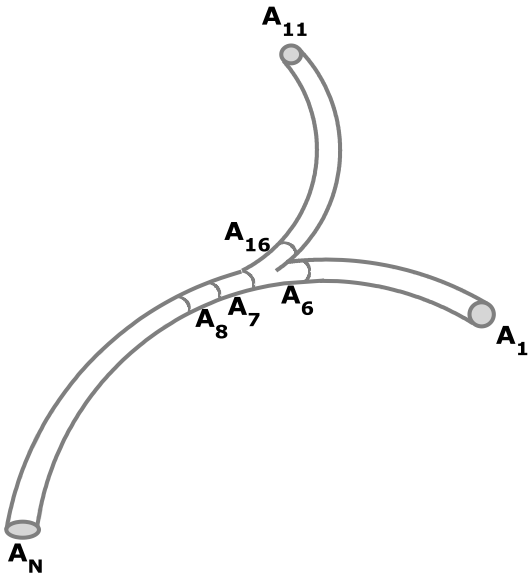

**Fig. 2 Merging dendrites ($A_{11}$ and $A_1$ are Boolean inputs; $A_N$ is a Boolean output)**

But when both inputs $A_1$, $A_{11}$ are active, pulses collide in segment 7; they might annihilate but under certain conditions they surprisingly propagate. Conditions for this propagation may be discovered by varying $R_7$, the series resistance in segment 7, while other series resistances ($R$), are left at their nominal values, *99.9 M*. Above $R_7 \approx 200\ M$ one or more input pulses give no $A_N$ output; below $R_7 \approx 180\ M$ input pulses give the OR function at $A_N$. Only in the range *180 M < $R_7$ < 200 M* does the exclusive OR result at $A_N$:
$$A_N = A_1 \overline{A}_{11} + \overline{A}_1 A_{11}$$

The membrane is assumed to be active with no interference from inhibitory neurotransmitters or local myelination. If there were inhibitory neurotransmitters or local myelination, the current sources would be shut down. For example, if $i_{Na7}$ and $i_{K7}$ are removed, one may obtain an AND function (Burger 2008, 2009).

Why do two solitons annihilate to generate an XOR gate? The circuit model in Fig. 3 suggests that both segments 6 and 16 aid in pulling $V_7$ down. So for simultaneous pulsing a slightly narrow surge is expected in segment 7 (shown in Fig. 4), narrow enough to prevent a trigger in segment 8; a single incoming pulse experiences less pull-down, giving a wider pulse in segment 7 (shown in Fig. 5) and thus more charge to trigger segment 8.

**Conclusions**

For a fairly wide range of channel resistance within continuously active membrane, a dendritic vertex may compute a deterministic XOR or for that matter, any arbitrary logic. This logic is independent of: 1) distributed processing 2) inhibitory neurotransmitters 3) local myelination 4) concocted nonlinearities. However, it is required that pulses arrive simultaneously at the vertex. Conditions for dendritic vertex logic are easily demonstrated using circuit models for an artificial dendritic summit. In fact, arbitrary Boolean logic easily occurs in a single vertex without resort to concepts offhandedly borrowed from traditional artificial neural networks.

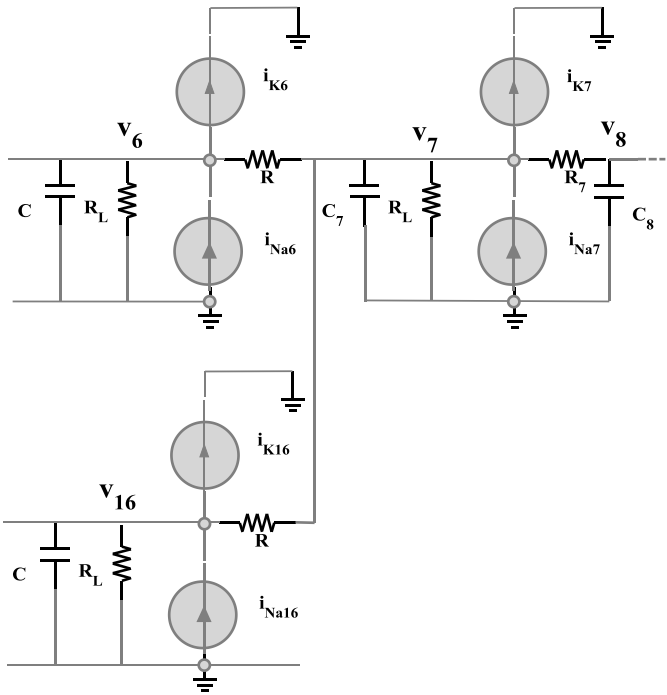

**Fig. 3.** Circuit model of merging active branches (inputs on left; output on right). This is a circuit for an artificial dendritic vertex.

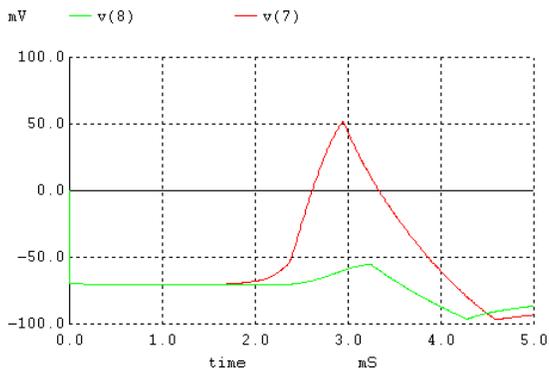

**Fig. 4.** XOR with both inputs active, showing a more narrow pulse and no triggering for $V_8$, shown as v(8) in this figure. $R_7 = 200M$ in this run.

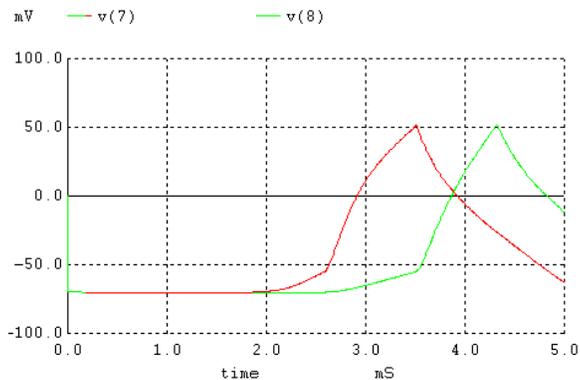

**Fig. 5.** XOR with one input active, showing a wider pulse and triggering for $V_8$, shown as v(8) in this figure. $R_7 = 200M$ in this run.

This may come as a surprise to engineers lost in a forest of circuits, but CMOS switches have fundamental limitations.  Coated conductors might someday be developed to give what dendrites give, which is the performance modeled by Fig. 3.  Advantages are:  1) efficiency, since there is controlled charging to minimize heat release, which is a problem in small artificial brains.  2) Cost effectiveness, since radically differing logic gates have a similar physical form, making them easier to manufacture.